\documentclass[runningheads]{llncs}

\usepackage{eccv}

\usepackage{eccvabbrv}

\usepackage{graphicx}
\usepackage{booktabs,makecell,multicol,multirow}
\usepackage{amsmath}
\usepackage{subcaption}
\usepackage{enumitem}
\usepackage[table]{xcolor}
\usepackage{verbatim}

\newcommand{\pos}[1]{{\hspace{0.3pt}\fontsize{4.5}{5}\selectfont\color{red!80!black}{(+$#1$)}}}
\newcommand{\negv}[1]{{\hspace{0.3pt}\fontsize{4.5}{5}\selectfont\color{blue!80!black}{(-$#1$)}}}
\newcommand{\neut}[1]{{\hspace{0.3pt}\fontsize{4.5}{5}\selectfont\color{gray}{(+$#1$)}}}

\usepackage[accsupp]{axessibility}  

\usepackage{hyperref}

\usepackage{orcidlink}

\begin{document}

\title{Towards Memory-Efficient Autoregressive Video Generation via Instance-Specific Parametric Absorption} 

\titlerunning{Memory-Efficient Video Generation by Parametric Absorption}

\author{
Xiaomeng Fu\inst{1,3}$^{\S,*}$\orcidlink{0000-0001-7195-0765} 
\and Jia Li\inst{2}$^\S$ 
\and Yiming Hu\inst{3}$^{\ddagger}$
\and Yong Wang\inst{3}$^{\dagger,\ddagger}$
\and Hayden Kwok-Hay So\inst{2}$^{\dagger}$
\and Jiao Dai\inst{1}$^{\dagger}$\orcidlink{0000-0003-3559-8009}
\and Xiangxiang Chu\inst{3}
\and Jizhong Han\inst{1}\orcidlink{0000-0003-1107-3873}
}

\authorrunning{X. Fu et al.}

\institute{Institute of Information Engineering, Chinese Academy of Sciences 
\and The University of Hong Kong
\and AMAP, Alibaba Group
}

\maketitle

\begingroup
\makeatletter
\renewcommand{\thefootnote}{}
\long\def\@makefntext#1{\noindent#1}
\footnotetext{\mbox{$^\S$ Equal contribution\hspace{1.2em}$^{\ddagger}$ Project lead\hspace{1.2em}$^{\dagger}$ Corresponding author.}\\
* This work was done during an internship at AMAP, Alibaba Group.
}
\makeatother
\endgroup

\begin{abstract}
    Autoregressive (AR) streaming models have emerged as a powerful paradigm for long video generation. However, the linearly growing Key-Value (KV) cache poses a significant bottleneck, leading to memory overload and degraded inference throughput. A common compression method is to drop redundant KV tokens, which often breaks long-range dependencies, resulting in temporal flickering and identity loss. In this paper, we propose Instance-Specific Parametric Absorption (ISPA), a novel framework that shifts the KV cache compression from discarding to distilling. The core idea is to transit a subset of layers from Full-Attention (F-Layers) to memory-efficient Local-Attention (L-Layers) by "absorbing" historical context into the model's weights. Specifically, during a brief warmup phase, ISPA monitors the output discrepancy between global and local attention. At the transition point, we solve a closed-form least-squares problem to compute an instance-specific weight modulation that compensates for the missing history. Experiments across architectures (1.3B to 14B) demonstrate that ISPA can remove up to 50\% of the KV cache with near-lossless visual quality. We hope this perspective encourages future work to explore parametric memory consolidation beyond external token-level cache management for streaming generative models.
    
\end{abstract}

\section{Introduction}
Diffusion models have become a central paradigm in visual generation, powering diverse tasks~\cite{fu2025tcfg,li2026rethinking}. Among these tasks, video generation~\cite{ma2024latte,wan2025wan,polyak2024movie,villegasphenaki,blattmann2023stable,yang2024cogvideox} is particularly challenging, as it requires both high visual fidelity and long-range temporal coherence. Recent works extend diffusion~\cite{ho2020denoising,rombach2022high} and flow-matching models~\cite{lipmanflow,ma2024sit,liuflow,esser2024scaling,peebles2023scalable} from image-level generation to video synthesis, typically by generating a fixed-length clip holistically ~\cite{ma2024latte,wan2025wan,polyak2024movie,villegasphenaki,blattmann2023stable,yang2024cogvideox}. While effective for short videos, this paradigm requires full spatiotemporal attention across all frames, whose quadratic memory and computation costs make long or infinite video generation prohibitively expensive.

To address this limitation, the Autoregressive (AR) streaming paradigm~\cite{yin2025slow,huang2025self,yang2025longlive,krea_realtime_14b,huang2025live,lu2025reward,krea_realtime_14b,teng2025magi,huang2026live,chen2025skyreels} has become a popular solution. By breaking a video into sequential segments, AR models generate frames one by one, using a Key-Value (KV) cache to store historical context. This method theoretically allows for videos of any length and supports real-time generation. However, the streaming process has a major drawback: to keep the video consistent over time, the KV cache must store continually accumulated history. As generation continues, this memory bank fills the GPU memory and significantly slows down the inference speed, losing the "real-time" advantage of the system.

A common way to reduce this memory pressure is to drop redundant tokens from the KV cache. While token dropping has been effective for Large Language Models (LLMs), directly applying this idea to dynamic video signals is often unstable. The key issue is that, in long video generation, the KV cache is not merely a storage buffer, but an external token memory that carries the model’s accumulated visual history. Existing pruning or eviction strategies manage this memory outside the model by deciding which tokens to keep or discard. Once important historical tokens are removed, long-range dependencies can be broken, causing the model to forget the past and leading to background flickering, structural distortion, and identity loss.

These challenges lead us to rethink how context should be compressed in a streaming setting. Instead of treating the KV cache as an ever-growing external memory to be repeatedly edited, we ask whether part of this memory can be consolidated into the model itself. The streaming nature of AR generation provides exactly such an opportunity: as a video unfolds, its instance-specific temporal patterns become observable. In some layers, the accumulated history then behaves as a stable contextual bias for future attention outputs, rather than as fully dynamic token-level information. Such history does not necessarily need to remain as explicit KV tokens; it can instead be analytically absorbed into the model’s linear weights through instance-specific test-time adaptation.

Based on this insight, we propose Instance-Specific Parametric Absorption (ISPA), a novel framework that reframes KV cache compression as parametric memory consolidation. ISPA transitions selected layers from Full-Attention Layers (F-Layers) to memory-efficient Local-Attention Layers (L-Layers) during inference. In a standard F-Layer, the model must query the growing KV cache to maintain temporal consistency. In contrast, an L-Layer attends only to local frames, which is efficient but risks losing historical context. Instead of simply dropping the missing history, ISPA performs Layer Absorption: it treats the difference between full and local attention as a reconstruction problem and solves for an instance-specific weight modulation in closed form. The resulting L-Layers can emulate the effect of long-range history using only local information, effectively internalizing part of the external KV memory into the model parameters.

A major technical challenge is performing this adaptation without interrupting the real-time generation process. To eliminate the computational overhead of monitoring the necessary attention signals (both local and full attention), we design a Decomposable Attention mechanism. By leveraging the Log-Sum-Exp (LSE) states already tracked by hardware-aware kernels (e.g., FlashAttention), we can collect both local and global attention during the standard forward pass. This enables ISPA to compute the weight updates without gradient-based optimization, thereby preserving the original inference latency of the base model.

We evaluate ISPA across several state-of-the-art architectures~\cite{krea_realtime_14b,huang2025live,lu2025reward,krea_realtime_14b} ranging from 1.3B to 14B parameters and various tasks (text-to-vieo and speech-to-video). ISPA successfully removes up to 50\% of the KV cache with near-lossless (<1\%) visual quality. Moreover, we demonstrate that ISPA is intrinsically compatible with post-training quantization. By combining parametric absorption with W8A8 quantization, we achieve an aggregate 1.86$\times$ inference speedup. 
Our contributions are three-fold:
\begin{itemize}[leftmargin=*,itemsep=2pt,topsep=0pt,parsep=0pt,label=$\bullet$]
    \item We introduce ISPA, a novel streaming framework that shifts the KV cache reduction from "pruning" to "absorption". Instead of discarding historical context, ISPA distills instance-specific context into the model's weights to maintain global coherence while operating with a minimal local cache.
    \item We propose a non-iterative, closed-form optimization for weight modulation. By leveraging a Decomposable Attention mechanism grounded in Online Softmax, our method collects necessary signals and updates parameters during the standard forward pass with negligible overhead. 
    \item We conduct extensive experiments on ISPA across various AR models (from 1.3B to 14B, from text-to-video to speech-to-video). Our results demonstrate that ISPA can remove up to 50\% of the KV cache with near-lossless visual quality and an aggregate 1.86$\times$ speedup when combined with quantization.
\end{itemize}
\section{Preliminaries}
\noindent \textbf{Video Diffusion Models.}
Most modern video generators~\cite{wan2025wan,blattmann2023stable} are built by extending diffusion or flow models from the image domain to the video domain. A video clip is typically represented as a 4D tensor $\mathbf{X}\in\mathbb{R}^{F\times C\times H\times W}$, where $F$ is the number of frames. Generation originates from Gaussian noise, gradually transforming into a clean video through an iterative denoising process. In Rectified Flow, the model predicts a velocity at each step that transports the current sample towards the target data distribution. 

A central challenge in video generation is maintaining temporal coherence: frames must remain consistent over time. The standard solution is to process all frames jointly utilizing spatiotemporal self-attention, allowing tokens from any frame to attend to tokens from any other frame. While highly effective, this operation is notoriously expensive; attention across $F$ frames typically scales as $\mathcal{O}(F^2)$ in both time and memory footprint, rendering long-video generation computationally prohibitive.

\noindent \textbf{Autoregressive Video Generation.}
To scale to longer sequence lengths, an autoregressive (AR) strategy~\cite{huang2025self,yin2025slow,yang2025longlive} generates the video in smaller, consecutive chunks. Let a long video $\mathbf{X}$ be partitioned into $N$ chunks:
\begin{equation}
    \mathbf{X}=\{\mathbf{x}^{(1)},\mathbf{x}^{(2)},\dots,\mathbf{x}^{(N)}\}
\end{equation}
where each chunk contains $L$ frames. The AR factorization is defined as:
\begin{equation}
p(\mathbf{X})=p(\mathbf{x}^{(1:N)})=\prod_{n=1}^{N}p(\mathbf{x}^{(n)}\mid \mathbf{x}^{(<n)})
\end{equation}
Each chunk $\mathbf{x}^{(n)}$ is generated conditioned on the previously generated history $\mathbf{x}^{(<n)}$. By maintaining a sliding window of past context, AR models can theoretically extrapolate to videos of arbitrary length, effectively shifting the bottleneck from hardware memory limits to sequential inference latency.

\noindent \textbf{Conditioning on the Past via KV Injection.}
A defining design choice in AR video diffusion is how the current chunk's denoiser leverages historical information. A prevalent approach is conditioning via cross-chunk self-attention: at denoising step $t$ for chunk $n$, the model derives queries, keys, and values from the current noisy chunk $\mathbf{x}^{(n)}_t$. To incorporate the past, the keys and values from previously generated frames are concatenated to the current chunk's respective tensors:
\begin{equation}
    \mathbf{Q}=\mathbf{Q}_{\text{cur}},\quad
    \mathbf{K}=[\mathbf{K}_{\text{past}};\mathbf{K}_{\text{cur}}],\quad
    \mathbf{V}=[\mathbf{V}_{\text{past}};\mathbf{V}_{\text{cur}}]
\end{equation}
Here, $(\mathbf{Q}_{\text{cur}},\mathbf{K}_{\text{cur}},\mathbf{V}_{\text{cur}})$ originate from the current chunk $\mathbf{x}^{(n)}_t$, while $(\mathbf{K}_{\text{past}},\mathbf{V}_{\text{past}})$ are derived from the generated history $\mathbf{x}^{(<n)}$. Specifically, $(\mathbf{K}_{\text{past}},\mathbf{V}_{\text{past}})$ are typically truncated to include only the KV pairs from the most recent frames to ensure memory efficiency. Intuitively, the current chunk "queries" a persistent memory bank of past frames. This KV-based injection is critical for both visual quality and generation efficiency, as it determines how temporal dependencies propagate through the autoregressive diffusion process.

\begin{figure*}[t]
    \centering
    \includegraphics[width=\linewidth]{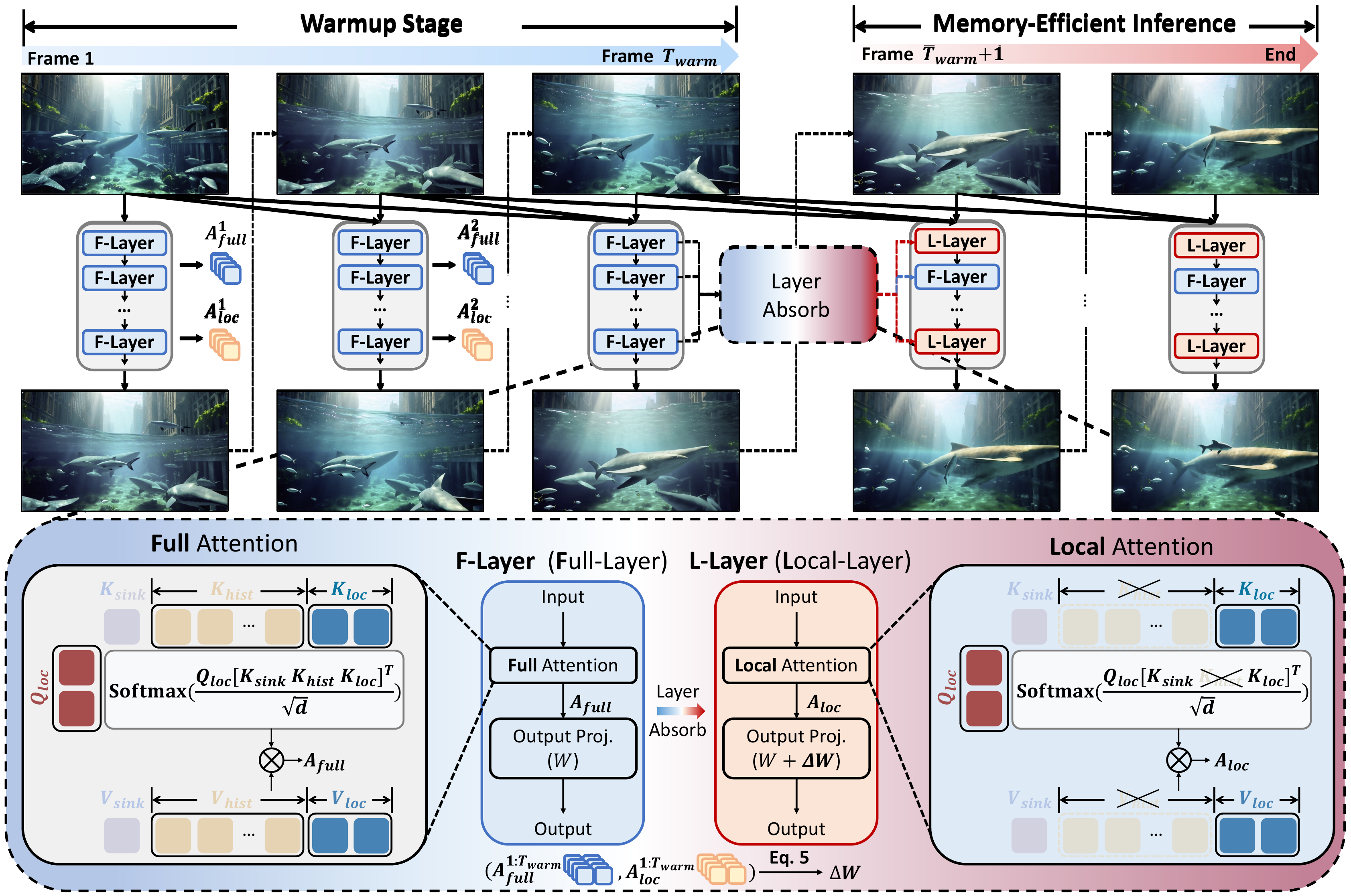} 
    \caption{\textbf{The overall framework of ISPA.} ISPA consists of three phases: (1) \textbf{Dual-Stream Warmup}: During the first $T_{warm}$ frames, F-Layers collect full attention outputs $A_{full}$ and local outputs $A_{loc}$. (2) \textbf{Layer Absorb}: At frame $T_{warm}$, we select $K$ layers and convert them to L-Layer (Local Layer). (3) \textbf{Memory-Efficient Inference}: The selected layers permanently evict historical KV caches and utilize $W+\Delta W$ to compensate for the lost context, while the remaining $N-K$ layers stay as F-Layers.}
    \label{fig:framework}
\end{figure*}

\section{Method}
To achieve exhaustive instance-level KV cache compression without compromising real-time throughput, we propose \textbf{I}nstance-\textbf{S}pecific \textbf{P}arametric \textbf{A}bsorption (\textbf{ISPA}), a streaming compression framework. Our goal is to mine the inherent temporal redundancy within each video generation instance and compress it, while maintaining strictly negligible compute overhead to preserve the realtime property of autoregressive inference. 

\subsection{Overall Framework}
To circumvent the memory explosion of the KV cache, ISPA shifts from conventional token pruning which irreversibly discards historical data to "Absorption" paradigm. This paradigm distills historical context into the model’s weights via \textit{Instance-Specific} test-time adaptation, overfitting parameters to the unique temporal dynamics of the specific video sequence currently being generated.
As illustrated in Figure~\ref{fig:framework}, ISPA operates through a streamlined, three-stage pipeline: 
\begin{enumerate}[leftmargin=*, align=left, labelsep=0.5em]
    \item \textbf{Dual-Stream Warmup:} During the generation of the initial $T_{warm}$ frames, the model operates in a dual-stream configuration using \textbf{F-Layers} (Full-Layers). We simultaneously collect the "Full" attention activation $\mathbf{A}_{full}$ (utilizing the whole historical context) and the "Local" activation $\mathbf{A}_{loc}$ (utilizing only current and sink tokens) via an optimized kernel-level decomposition.
    \item \textbf{Layer Absorb:} At the dividing frame $T_{warm}$, the accumulated dual-stream activations serve as the basis for weight adaption. We formulate the absorption as a reconstruction problem, computing the optimal weight modulation $\Delta W$ for the output projection to bridge the gap between local and full attention. Rather than treating all layers equally, we employ a dynamic ranking mechanism to identify $K$ "absorbable" layers that most accurately reconstruct global dependencies through weight modulation.
    \item \textbf{Memory-Efficient Inference:} For all subsequent frames $t > T_{warm}$, the designated $K$ layers transition into \textbf{L-Layers} (Local-Layers). These layers completely discard their historical KV cache ($\mathcal{K}_{hist}, \mathcal{V}_{hist}$), retaining only the local context. The absence of this history is effectively compensated for by the newly updated weights $W_{new}=W+\Delta W$. The remaining $N-K$ layers continue as F-Layers to maintain global coherence.
\end{enumerate}

Although we present this warmup-to-absorption transition as a one-shot pipeline for clarity, the same procedure can be re-triggered when the video enters a new temporal regime, allowing ISPA to recalibrate the absorbed layers and their weight modulation for the new context.

\subsection{Layer Absorb via Full Attention Reconstruction}
The computational core of ISPA lies in absorbing the historical context of a layer into its weights. We formulate this as a unified least-squares optimization problem, solved at the end of the warmup phase.

As shown in the "F-Layer" block of Figure~\ref{fig:framework}, for each diffusion transformer layer $l\in\{1,2,...,N\}$, we monitor the internal activations across $T_{warm}$ frames. Let $\mathbf{A}_{loc,l}^{1:T_{warm}} \in \mathbb{R}^{T_{warm}\times D}$ denote the attention output immediately preceding the final linear projection layer $W$, computed utilizing \textit{only} the local context (current frame and sink frames). Let $\mathbf{A}_{full,l}^{1:T_{warm}} \in \mathbb{R}^{T_{warm}\times D}$ represent the ground-truth "full" output of that same attention block. \textbf{For notational brevity, we omit the layer index and the temporal superscripts in the following derivation.} Our objective is to discover a weight modulation matrix $\Delta W$ such that the local-stream output after projection matches the full-stream output. Specifically, we target the linear projection layer for adaption as it is the immediate downstream component of the attention block. By compensating for the context loss at this point of origin, we minimize the propagation of reconstruction errors to subsequent layers. The optimization is formulated as:
\begin{equation}
\min_{\Delta W} || \mathbf{A}_{loc} (W + \Delta W) - \mathbf{A}_{full} W ||^2_2 + \lambda || \Delta W ||^2_2
\end{equation}
By defining the target residual signal as $\mathbf{R} = (\mathbf{A}_{full} - \mathbf{A}_{loc})W$, the optimal $\Delta W$ that absorbs the truncated history can be derived via the closed-form regression solution:
\begin{equation}
\Delta W = ( (\mathbf{A}_{loc})^\top \mathbf{A}_{loc} + \lambda \mathbf{I} )^{-1} (\mathbf{A}_{loc})^\top \mathbf{R}
\label{eq:close_form_adapt}
\end{equation}
The reconstruction error (feasibility score) for each layer is then defined as the Frobenius norm of the residual:
\begin{equation}
    \epsilon = || \mathbf{A}_{loc} (W + \Delta W) - \mathbf{A}_{full} W ||^2_F
\end{equation}
A major advantage of this formulation is its strictly \textbf{non-iterative} nature. Unlike traditional knowledge distillation that necessitates backpropagating gradients through massive computational graphs, ISPA only requires summing the outer products of activations ($\mathbf{A}_{loc}^\top \mathbf{A}_{loc}$ and $\mathbf{A}_{loc}^\top \mathbf{R}$) during the standard forward pass. This reduces optimization to a single $D\times D$ matrix inversion precisely at frame $T_{warm}$. Consequently, the adaptation signal is harvested and applied without ever initializing a gradient-based optimizer, rigorously preserving the real-time inference pipeline.

\subsection{Dynamic Layer Selection}
Having computed the candidate update $\Delta W$ and error metric $\epsilon$ for all layers at the end of $T_{warm}$, we dynamically route which layers will transition to a KV cache-free regime.

Rather than enforcing a fixed subset of layers across all prompts, ISPA executes Instance-Specific Selection. Given a hardware budget of $K$ layers designated for KV cache eviction, we rank layers in ascending order based on $\epsilon$. The $K$ layers exhibiting the lowest $\epsilon$ are designated as \textbf{L-Layers}. For these layers, we inject the update $W_{new}=W+\Delta W$ and \textbf{permanently} evict their historical KV cache $\mathcal{K}_{hist}$. The remaining $N-K$ layers, which evidently handle complex temporal dependencies that resist linear parametric absorption, remain as \textbf{F-Layers} to guarantee generation quality.

This dynamic routing is essential because diverse video sequences exhibit significantly different temporal dynamics and these dynamics heterogeneously distribute across the model's hierarchy. 
By delaying the architectural decision until $T_{warm}$, ISPA autonomously tailors the model's topology to the specific instance, a phenomenon we term test-time structural elasticity. Once the topology is locked, the model's parameters are updated in-place. For all subsequent frames $t>T_{warm}$, L-Layers execute using only localized KV pairs, yet their modified $W_{new}$ simulates the continuous presence of the evicted history.

Importantly, this calibration is not permanently tied to the initial scene. The one-shot schedule described above is the default setting used in our experiments, but the same procedure can be re-triggered when the video undergoes a large scene transition or temporal distribution shift. In such cases, ISPA briefly re-enters the dual-stream warmup stage, recomputes the layer-wise reconstruction errors, and refreshes both the absorbed layer set and the corresponding $\Delta W$ for the new context. Since this adaptation is closed-form and lightweight, event-triggered recalibration preserves the streaming nature of inference while avoiding stale weight modulation under changing scenes.

\subsection{Real-time Efficiency Optimization via Decomposable Attention}
A potential bottleneck of ISPA is the necessity to simultaneously collect two distinct attention signals (the localized attention $\mathbf{A}_{loc}$ for $\Delta W$ computation and the full attention $\mathbf{A}_{full}$ for standard inference) during the warmup phase. A naive implementation would dual-process the attention mechanism, degrading real-time performance. To eliminate this overhead, we introduce a \textit{Decomposable Attention} mechanism grounded in the Online Softmax theorem.

For a video sequence at step $t\in [1, T_{warm}]$, we define the \textbf{Local Context} as the union of the initial frame (serving as the "sink frame", $K_{sink}=K_1$) and the current frame's KV ($K_{loc}=K_t$). Formally, the key and value sets are partitioned as follows:
\begin{align}
\mathcal{K}_{loc} = \{K_1, K_t\},& \quad \mathcal{V}_{loc} = \{V_1, V_t\} \\
\mathcal{K}_{hist} = \{K_2, \dots, K_{t-1}\},& \quad \mathcal{V}_{hist} = \{V_2, \dots, V_{t-1}\}
\end{align}
The exhaustive KV cache is the disjoint union: $\mathcal{K}_{full}=\mathcal{K}_{loc\_total} \cup \mathcal{K}_{hist}$.

By exploiting the Log-Sum-Exp (LSE) state inherently tracked by modern hardware-aware kernels~\cite{dao2022flashattention,daoflashattention,zhangsageattention,zhang2025sageattention2} (e.g., FlashAttention), we decouple the attention into two independent branches and fuse them post-hoc. For each layer $l$, we compute the local and historical outputs along with their corresponding LSE values:
\begin{align}
(\mathbf{O}_{loc}, LSE_{loc}) &= \text{FlashAttn}(\mathbf{Q}_t, \mathcal{K}_{loc}, \mathcal{V}_{loc}) \\
(\mathbf{O}_{hist}, LSE_{hist}) &= \text{FlashAttn}(\mathbf{Q}_t, \mathcal{K}_{hist}, \mathcal{V}_{hist})
\end{align}
$\mathbf{O}_{loc}$ is mathematically identical to $\mathbf{A}_{loc}$. To perfectly reconstruct the Full output $\mathbf{A}_{full}$ without recomputing attention, we apply a weighted fusion driven by the LSE values:
\begin{gather}
    LSE_{full} = \log \left( \exp(LSE_{loc}) + \exp(LSE_{hist}) \right)\\
    \mathbf{A}_{full} = \exp(LSE_{loc} - LSE_{full}) \cdot \mathbf{O}_{loc} + \exp(LSE_{hist} - LSE_{full}) \cdot \mathbf{O}_{hist}
\end{gather}
This dual-path decomposition ensures that the warmup phase is nearly "free" in terms of FLOPs. The total dot-product operations remain almost identical to a standard attention pass as the query vector is simply partitioned across the same total number of KV pairs. Since the fusion involves only scalar operations in SRAM, we avoid materializing massive intermediate attention matrices, preserving the memory efficiency of FlashAttention. The final fusion accounts for less than 1\% of execution time, ensuring a smooth transition from full-cache inference to parametric-absorbed inference.

\section{Experiments}

\begin{table*}[t!]
\centering
\caption{Quantitative evaluation on the VBench \textbf{5s} benchmarks (prompts from MovieGen). \textbf{Aes.}: aesthetic quality; \textbf{Back.}: background consistency; \textbf{Dyn.}: dynamic degree; \textbf{Image.}: imaging quality; \textbf{Mot.}: motion smoothness; \textbf{Sub.}: subject consistency; \textbf{Temp.}: temporal flickering. \textbf{ISPA-}$x$ indicates that a fraction $x$ of F-Layers are replaced by L-Layers (i.e., $K=xN$).}
\label{tab:main_results_5s}
\small
\renewcommand{\arraystretch}{1.3}
\setlength{\tabcolsep}{1.2pt} 

\newcommand{\phm}{{\hspace{0.3pt}\fontsize{4.5}{5}\selectfont\phantom{(+0.00)}}}

\resizebox{\textwidth}{!}{
\begin{tabular}{ll c lllllll} 
\toprule
\multirow{2}{*}{\textbf{Model}} & \multirow{2}{*}{\textbf{Method}} & \multirow{2}{*}{\textbf{KVCache}} & \multicolumn{7}{c}{\textbf{VBench Metrics}} \\
\cmidrule(l){4-10}
& & & \textbf{Aes.$\uparrow$} & \textbf{Back.$\uparrow$} & \textbf{Dyn.$\uparrow$} & \textbf{Image.$\uparrow$} & \textbf{Mot.$\uparrow$} & \textbf{Sub.$\uparrow$} & \textbf{Temp.$\uparrow$} \\ 
\midrule
\multirow{4}{*}{\makecell[c]{LongLive \\ (1.3B)}} 
& Vanilla  & 100\% & 64.43\phm & 95.76\phm & 40.77\phm & 70.74\phm & 98.92\phm & 95.55\phm & 97.87\phm \\ 
& ISPA-0.3 & 72.5\% & 64.48\pos{0.05} & 95.80\pos{0.04} & 40.60\negv{0.17} & 70.64\negv{0.10} & 98.91\negv{0.01} & 95.53\negv{0.02} & 97.87\neut{0.00} \\
& ISPA-0.4 & 63.3\% & 64.47\pos{0.04} & 95.80\pos{0.04} & 40.60\negv{0.17} & 70.64\negv{0.10} & 98.91\negv{0.01} & 95.53\negv{0.02} & 97.87\neut{0.00} \\
& ISPA-0.5 & 54.2\% & 64.48\pos{0.05} & 95.78\pos{0.02} & 40.60\negv{0.17} & 70.70\negv{0.04} & 98.92\neut{0.00} & 95.59\pos{0.04} & 97.86\negv{0.01} \\ 
\midrule
\multirow{4}{*}{\makecell[c]{Reward \\ (1.3B)}} 
& Vanilla  & 100\% & 62.82\phm & 94.73\phm & 68.96\phm & 68.65\phm & 98.08\phm & 93.76\phm & 96.34\phm \\ 
& ISPA-0.3 & 73.3\% & 62.83\pos{0.01} & 94.74\pos{0.01} & 69.50\pos{0.54} & 68.65\neut{0.00} & 98.08\neut{0.00} & 93.76\neut{0.00} & 96.34\neut{0.00} \\
& ISPA-0.4 & 64.4\% & 62.82\neut{0.00} & 94.74\pos{0.01} & 69.17\pos{0.21} & 68.65\neut{0.00} & 98.08\neut{0.00} & 93.76\neut{0.00} & 96.34\neut{0.00} \\
& ISPA-0.5 & 55.6\% & 62.85\pos{0.03} & 94.73\neut{0.00} & 69.60\pos{0.64} & 68.65\neut{0.00} & 98.08\neut{0.00} & 93.76\neut{0.00} & 96.34\neut{0.00} \\ 
\midrule
\multirow{4}{*}{\makecell[c]{Krea \\ (14B)}} 
& Vanilla  & 100\% & 64.50\phm & 95.40\phm & 77.12\phm & 70.86\phm & 98.62\phm & 94.90\phm & 96.46\phm \\ 
& ISPA-0.3 & 75.0\% & 64.51\pos{0.01} & 95.16\negv{0.24} & 76.65\negv{0.47} & 71.03\pos{0.17} & 98.71\pos{0.09} & 94.66\negv{0.24} & 96.91\pos{0.45} \\
& ISPA-0.4 & 66.7\% & 64.66\pos{0.16} & 94.68\negv{0.72} & 76.15\negv{0.97} & 70.70\negv{0.16} & 98.80\pos{0.18} & 94.02\negv{0.88} & 97.14\pos{0.68} \\
& ISPA-0.5 & 58.3\% & 64.93\pos{0.43} & 94.06\negv{1.34} & 76.23\negv{0.89} & 71.01\pos{0.15} & 98.72\pos{0.10} & 92.32\negv{2.58} & 97.04\pos{0.58} \\ 
\bottomrule
\end{tabular}
}
\end{table*}

\subsection{Experimental Settings}
\noindent \textbf{Models and Benchmarks.}
To evaluate the generalizability and robustness of our proposed ISPA, we conduct extensive experiments across four representative autoregressive video generation models: \textbf{LongLive}~\cite{yang2025longlive}, \textbf{Reward-Forcing}~\cite{lu2025reward} (denoted as \textbf{Reward}), \textbf{Krea-Realtime}~\cite{krea_realtime_14b} (denoted as \textbf{Krea}), and \textbf{LiveAvatar}~\cite{huang2025live}. These models cover a diverse range of model size (from 1.3B to 14B) and generation task (including text-to-video and speech-to-video). The selection of these heterogeneous architectures, spanning from general-purpose synthesis to specialized human-centric generation, allows us to validate ISPA’s performance across varying modeling paradigms. For long video evaluation, we primarily employ the \textbf{VBench-Long} benchmark~\cite{huang2024vbench,zheng2025vbench}. In addition to the standard prompts provided by VBench, we incorporate complex full-length prompts from \textbf{MovieGen}~\cite{polyak2024movie} to further challenge the models' capability in maintaining long-range temporal consistency and semantic alignment. This combination ensures a comprehensive assessment that covers both basic motion quality and high-level narrative coherence. All ablation studies and analytical experiments are conducted based on the LongLive model as default backbone.

\begin{table*}[t!]
\centering
\caption{Quantitative evaluation on the VBench \textbf{30s} benchmarks (prompts from MovieGen). Notations and metric definitions follow Tab~\ref{tab:main_results_5s}.}
\label{tab:main_results_30s}
\small
\renewcommand{\arraystretch}{1.3}
\setlength{\tabcolsep}{1.2pt} 

\newcommand{\phm}{{\hspace{0.3pt}\fontsize{4.5}{5}\selectfont\phantom{(+0.00)}}}

\resizebox{\textwidth}{!}{
\begin{tabular}{ll c lllllll} 
\toprule
\multirow{2}{*}{\textbf{Model}} & \multirow{2}{*}{\textbf{Method}} & \multirow{2}{*}{\textbf{KVCache}} & \multicolumn{7}{c}{\textbf{VBench Metrics}} \\
\cmidrule(l){4-10}
& & & \textbf{Aes.$\uparrow$} & \textbf{Back.$\uparrow$} & \textbf{Dyn.$\uparrow$} & \textbf{Image.$\uparrow$} & \textbf{Mot.$\uparrow$} & \textbf{Sub.$\uparrow$} & \textbf{Temp.$\uparrow$} \\ 
\midrule
\multirow{4}{*}{\makecell[c]{LongLive \\ (1.3B)}} 
& Vanilla  & 100\% & 63.59\phm & 96.65\phm & 41.93\phm & 70.15\phm & 98.91\phm & 96.10\phm & 97.92\phm \\ 
& ISPA-0.3 & 72.5\% & 63.32\negv{0.27} & 96.48\negv{0.17} & 41.38\negv{0.55} & 69.38\negv{0.77} & 98.90\negv{0.01} & 95.77\negv{0.33} & 97.95\pos{0.03} \\
& ISPA-0.4 & 63.3\% & 63.05\negv{0.54} & 96.29\negv{0.36} & 42.37\pos{0.44} & 69.05\negv{1.10} & 98.88\negv{0.03} & 95.52\negv{0.58} & 97.90\negv{0.02} \\
& ISPA-0.5 & 54.2\% & 62.98\negv{0.61} & 96.28\negv{0.37} & 41.38\negv{0.55} & 69.05\negv{1.10} & 98.88\negv{0.03} & 95.61\negv{0.49} & 97.89\negv{0.03} \\ 
\midrule
\multirow{4}{*}{\makecell[c]{Reward \\ (1.3B)}} 
& Vanilla  & 100\% & 60.82\phm & 94.97\phm & 66.92\phm & 66.79\phm & 98.05\phm & 93.43\phm & 96.49\phm \\ 
& ISPA-0.3 & 73.3\% & 60.82\neut{0.00} & 94.98\pos{0.01} & 67.24\pos{0.32} & 66.79\neut{0.00} & 98.05\neut{0.00} & 93.43\neut{0.00} & 96.49\neut{0.00} \\
& ISPA-0.4 & 64.4\% & 60.72\negv{0.10} & 95.01\pos{0.04} & 65.45\negv{1.47} & 66.68\negv{0.11} & 98.09\pos{0.04} & 93.51\pos{0.08} & 96.58\pos{0.09} \\
& ISPA-0.5 & 55.6\% & 60.45\negv{0.37} & 94.95\negv{0.02} & 65.09\negv{1.83} & 66.55\negv{0.24} & 98.10\pos{0.05} & 93.36\negv{0.07} & 96.64\pos{0.15} \\ 
\midrule
\multirow{4}{*}{\makecell[c]{Krea \\ (14B)}} 
& Vanilla  & 100\% & 62.15\phm & 94.64\phm & 75.13\phm & 71.20\phm & 98.84\phm & 94.85\phm & 96.85\phm \\ 
& ISPA-0.3 & 75.0\% & 62.54\pos{0.39} & 95.32\pos{0.68} & 74.47\negv{0.66} & 70.46\negv{0.74} & 98.66\negv{0.18} & 94.91\pos{0.06} & 96.91\pos{0.06} \\
& ISPA-0.4 & 66.7\% & 63.29\pos{1.14} & 95.17\pos{0.53} & 74.80\negv{0.33} & 70.27\negv{0.93} & 98.81\negv{0.03} & 94.76\negv{0.09} & 97.26\pos{0.41} \\
& ISPA-0.5 & 58.3\% & 63.93\pos{1.78} & 94.63\negv{0.01} & 75.27\pos{0.14} & 70.12\negv{1.08} & 98.73\negv{0.11} & 93.81\negv{1.04} & 97.12\pos{0.27} \\ 
\bottomrule
\end{tabular}
}
\end{table*}

\noindent \textbf{Implementation Details.}
During the first 12 latent frames ($T_{warm}=12$), the model generates video using standard KV-Cache while simultaneously collecting $\mathbf{A}_{loc}$ and $\mathbf{A}_{full}$. At the switching point $T_{warm}$, we apply a \textbf{closed-form update} to the model parameters to absorb the historical context, after which the explicit KV-Cache is pruned. For the closed-form adaptation (Eq.~\ref{eq:close_form_adapt}), we employ a regularization parameter $\lambda=0.001$ by default to balance the preservation of pre-trained knowledge and the integration of new temporal context. 

\subsection{Main Results}
\noindent \textbf{Quantitative Results.} Tab~\ref{tab:main_results_5s} and~\ref{tab:main_results_30s} evaluates ISPA's ability to internalize historical context without suffering from "temporal amnesia". On standard 5s videos (1.3B models), ISPA achieves near lossless compression: evicting almost 50\% KV caches causes negligible deviations. Crucially, when scaling to 30s long-form generation, ISPA not only maintains minor degradation on 1.3B models but actively \textit{improves} Aesthetic Score and Temporal Consistency on the 14B Krea-Realtime model. We attribute this long-video improvement to our analytic absorption ($W_{new}=W+\Delta W$). Explicit attention over massive historical tokens often introduces high-frequency noise, corrupting diffusion trajectories. By forcing half of the layers to summarize context into static weights, ISPA explicitly filters this noise. While slightly blurring local token matching (minor Subject drop), it powerfully reinforces global semantic and temporal stability, acting as a macroscopic regularizer rather than just an efficiency optimization.

\begin{figure*}[t]
    \centering
    \includegraphics[width=\linewidth]{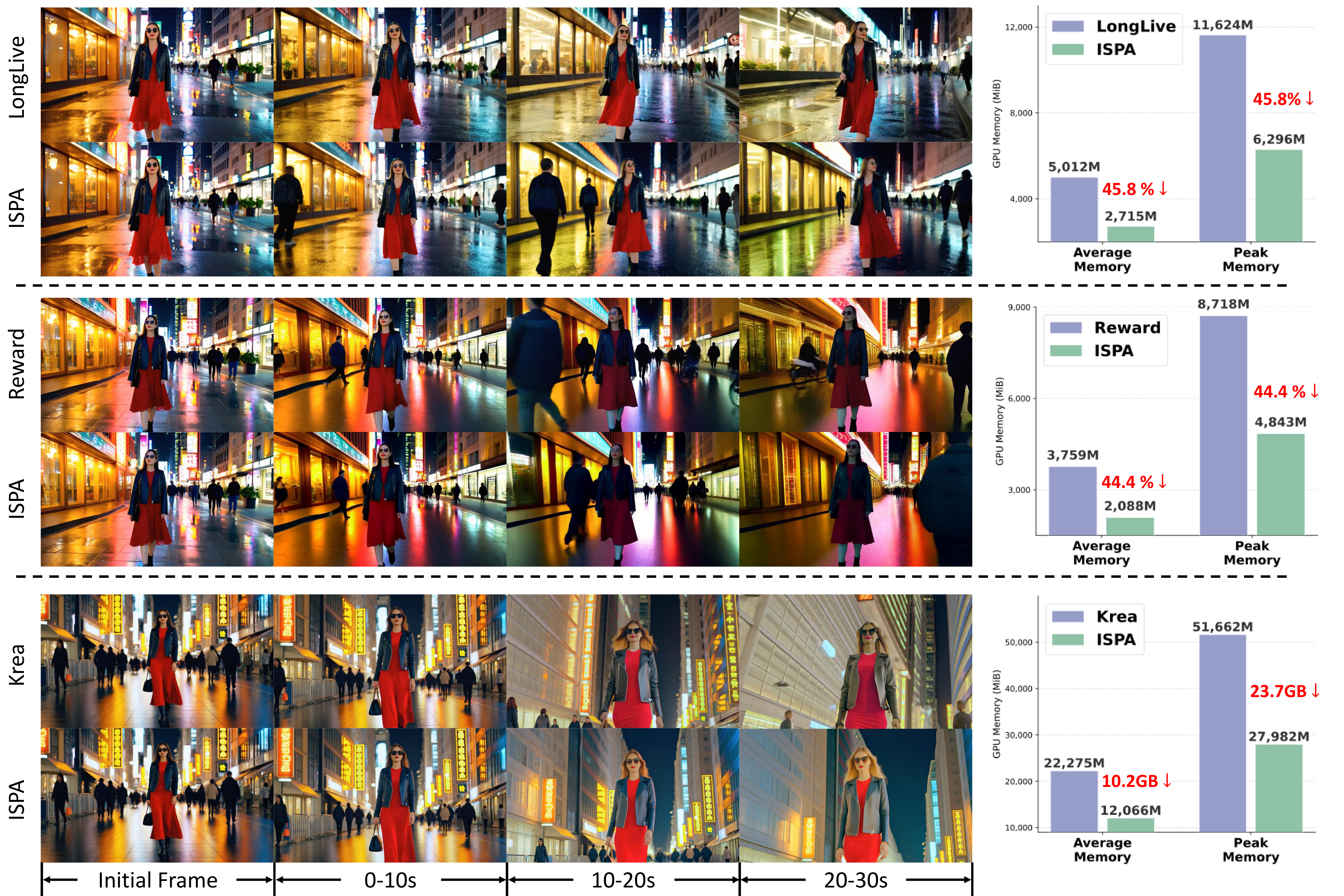} %
    \caption{Left: Qualitative comparison of 30-second long-form video generation. Right: GPU memory usage comparison.}
    \label{fig:qualitative_30s}
\end{figure*}

\noindent \textbf{Qualitative Results.}
Fig.~\ref{fig:qualitative_30s} visually validates ISPA's robustness in 30-second long-form generation. Across 1.3B to 14B models, ISPA aligns closely with full-cache baselines. Notably, at the 30-second mark (rightmost frames), ISPA consistently preserves intricate details. This confirms that our analytic absorption successfully distills historical context into projection weights, achieving substantial memory compression without visual degradation. Furthermore, the quantitative comparisons on the right panel highlight ISPA's superior memory efficiency. Across various architectures, ISPA consistently reduces GPU memory. Specifically, in the most Krea model (14B), ISPA achieves a remarkable reduction of 23.7GB in peak memory. This significant margin demonstrates that ISPA not only maintains high fidelity synthesis but also makes long-video generation feasible on consumer-grade hardware by effectively capping memory growth.

\begin{figure*}[t]
    \centering
    \includegraphics[width=\linewidth]{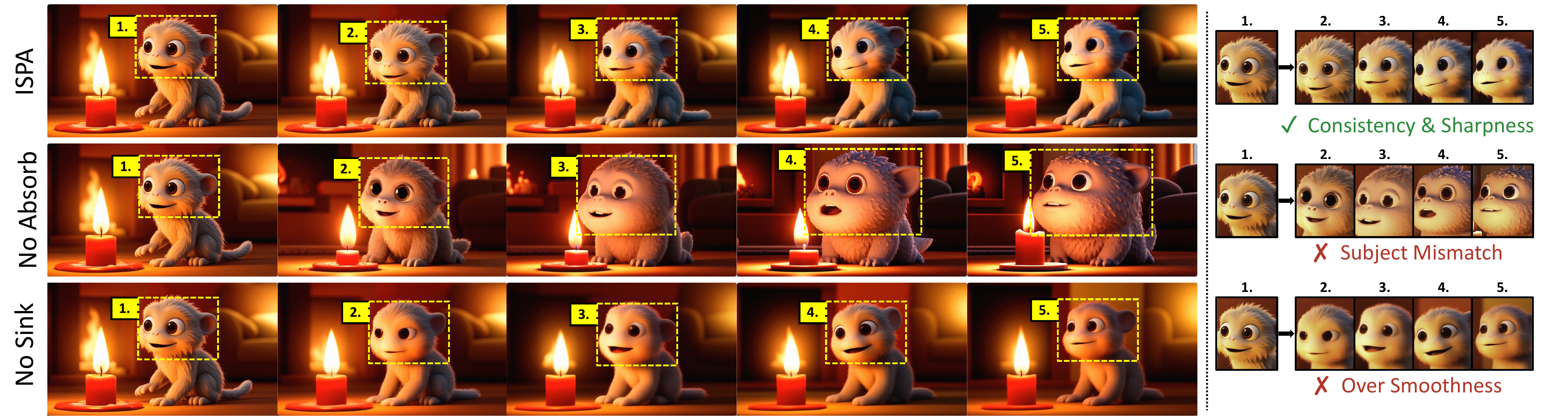} 
    \caption{Ablation study on core components. Top: Our full method maintains identity consistency and visual sharpness. Middle: Removing parametric adaptation (No Absorb) leads to subject morphing over time. Bottom: Omitting initial KV pairs (No Sink) causes oversmoothness.}
    \label{fig:ablation_absorb_sink}
\end{figure*}

\subsection{Ablation Study}
\noindent \textbf{The impact of Layer Absorb and Sink Frame.}
To analyze the mechanism of ISPA, we conduct an ablation study across two key dimensions: weight adaptation and attention anchoring (Fig.~\ref{fig:ablation_absorb_sink}). In the "No Absorb" setting (middle row), we evaluate the performance of simple local attention without our analytic $\Delta W$. This leads to a catastrophic "Subject Mismatch," where the creature's morphology drifts significantly as the generation progresses. This highlights that our instance-specific adaptation is not merely an auxiliary update, but a necessary operation to reconstruct the missing global receptive field. We also evaluate the performance with parametric adaptation while removing the sink frame ("No Sink", bottom row), which causes the model to generate "Over-Smoothed" results. We attribute this to the fact that the sink frame provides a stable basis for softmax normalization; its absence forces the attention mechanism to redistribute scores across less relevant local tokens, resulting in blurry, low-fidelity outputs. The combination between parametric absorption (for identity) and sink tokens (for stability) allows ISPA to match the quality of full-cache inference while maintaining a constant memory footprint.

\begin{figure}[t] %
    \centering
    \includegraphics[width=0.9\linewidth]{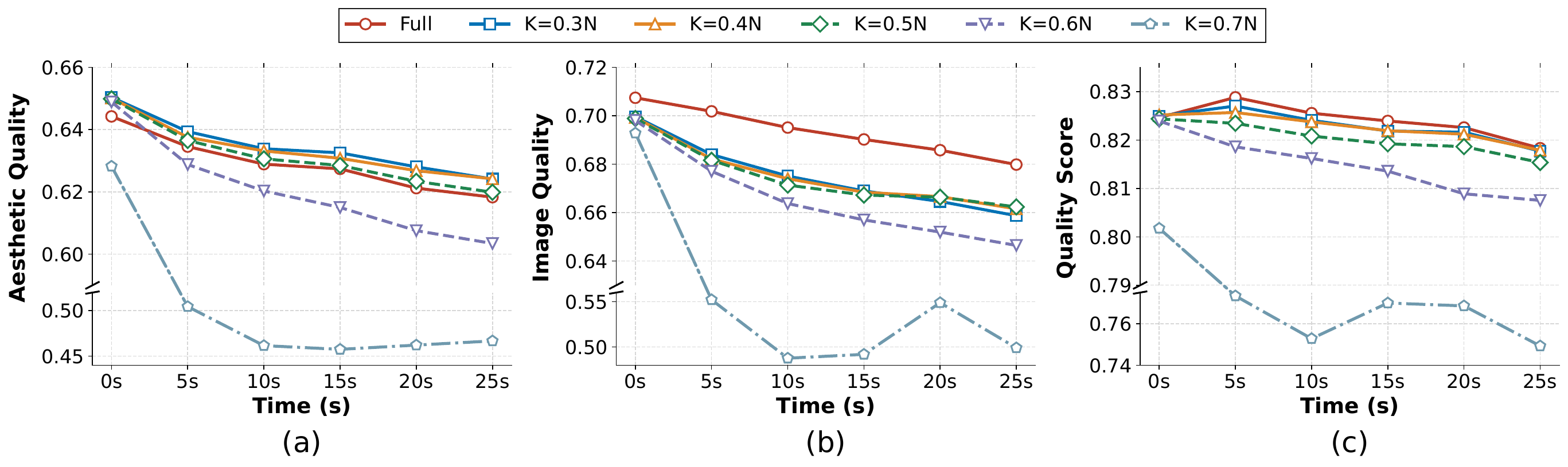}
    \caption{\textbf{Ablation on KV cache eviction ratios over time.} We evaluate (a) Aesthetic Quality, (b) Image Quality and Overall (c) Quality Score across a 30-second generation horizon.}
    \label{fig:absorb_layers_ablation}
\end{figure}

\noindent \textbf{The impact of the number of L-Layers $K$.} Figure~\ref{fig:absorb_layers_ablation} evaluates the robustness of ISPA across varying compression ratios (defined by the number of L-Layers $K$). For $K \le 0.5N$, ISPA maintains a performance nearly identical to the uncompressed baseline. Notably, in late-stage generation (>15s), ISPA with moderate $K$ even slightly outperforms the baseline in Aesthetic Quality. However, the video quality decreases as $K$ increase further (when $K=0.7N$). This reveals a efficiency-utility tradeoff: although linear absorption can preserve global context, a sufficient number of F-Layers is essential for modeling nonlinear, dynamic interactions. ISPA's dynamic selection mechanism effectively captures the tradeoff by preserving the most nonlinear layers, ensuring stability when the budge $K$ remains within a safe limit ($K\le 0.5N$). 

\begin{figure}[t] %
    \centering
    \includegraphics[width=0.9\linewidth]{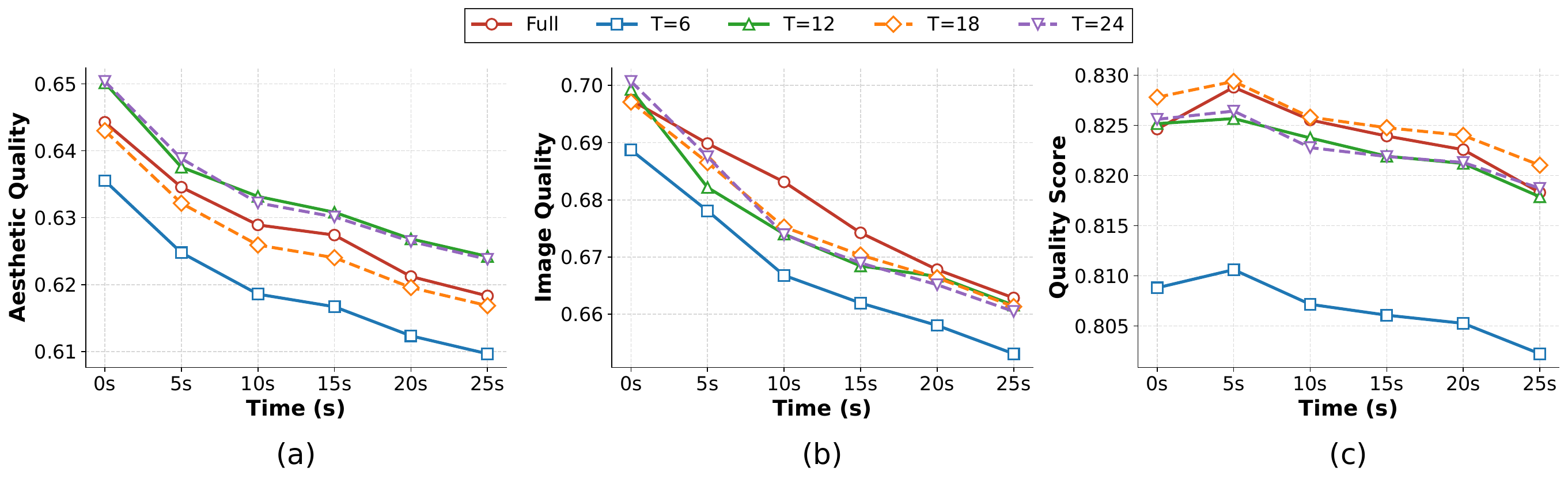} %
    \caption{\textbf{Impact of warm-up duration ($T_{warm}$).} We compare parametric adaptation using different initial frame counts. $T_{warm} \ge 12$ effectively prevents information starvation, matching or exceeding the full-cache baseline's long-term stability.}
    \label{fig:warmup_ablation}
\end{figure}

\noindent \textbf{The impact of warmup frame $T_{warm}$.} Figure~\ref{fig:warmup_ablation} investigates the sensitivity of our closed-form adaptation to the warmup frame. 
At $T_{warm}=6$, performance drops sharply across all metrics. 
We attribute this to \textbf{information starvation}: with insufficient temporal diversity, the derived $\Delta W$ fits to the narrow distribution of the first several frames and fails to generalize as the video's semantics evolve in later stages. 
Conversely, performance saturates and closely tracks the uncompressed baseline once $T_{warm} \ge 12$, confirming that a brief temporal window provides sufficient statistical evidence to capture the core temporal dynamics. 
While $T_{warm}=24$ remains stable, it delays the benefits of KV cache compression. 
Consequently, we adopt $T_{warm}=12$ as the optimal trade-off between adaptation accuracy and immediate memory efficiency.

\noindent \textbf{Impact of Warm-up Duration $T_{warmp}$.}
Figure~\ref{fig:warmup_ablation} evaluates the sensitivity of our closed-form adaptation to $T_{warmp}$. At T=6, performance degrades rapidly. We attribute this to \textit{information starvation}: lacking temporal diversity, the derived $\Delta W$ overfits to initial frames and fails to generalize to later semantic shifts. Conversely, for 
T$\ge$12, performance stabilizes and closely tracks the Full baseline, confirming that a brief window (2-3s) suffices to capture core temporal dynamics. Furthermore, $T\in [12,18]$ often outperforms the baseline in late stages, reinforcing that properly calibrated absorption filters long-range attention noise. Finally, while $T=24$ is stable, it delays KV cache eviction. We therefore adopt $T=12$ as the optimal trade-off, balancing mathematical convergence with immediate memory savings.

\begin{figure*}[t]
    \centering
    \includegraphics[width=\linewidth]{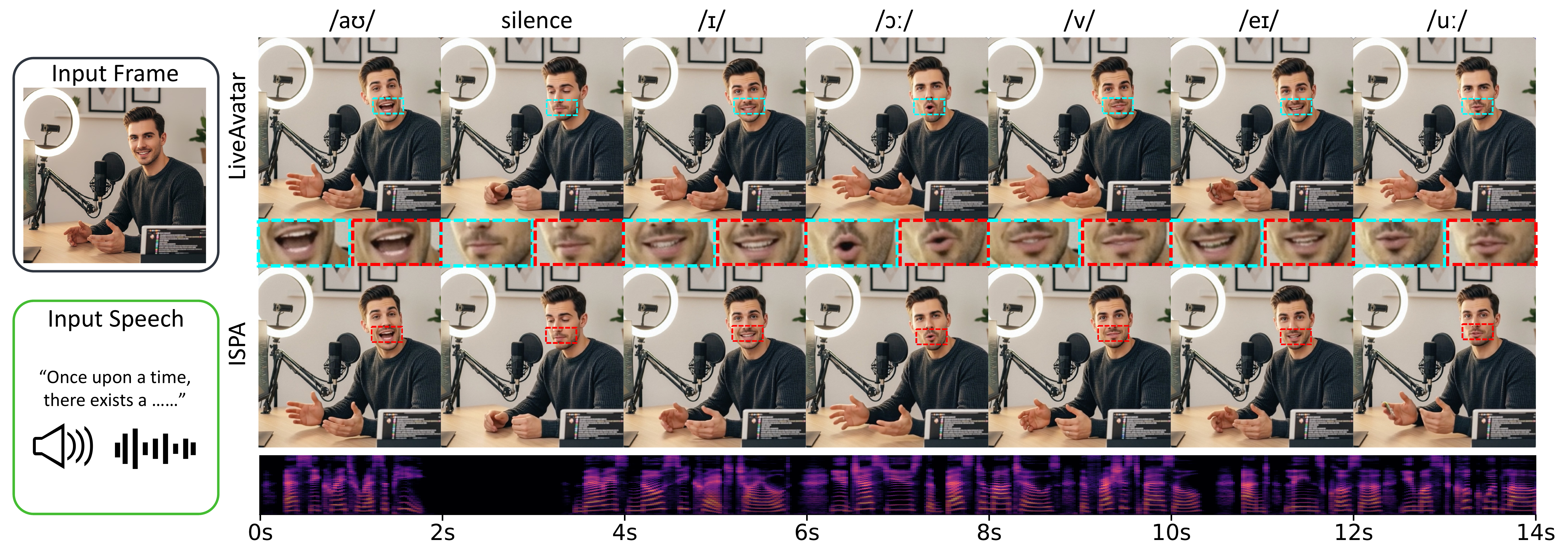}
    \caption{Generalization to Speech-to-Video (S2V) model. We integrate ISPA into the LiveAvatar for streaming talking-head synthesis. Given a single reference image and a continuous speech input, ISPA (bottom row) maintains nearly identical visual quality and viseme accuracy compared to the full-cache baseline (top row).}
    \label{fig:s2v_liveavatar}
\end{figure*}

\begin{figure*}[t]
    \centering
    \includegraphics[width=\linewidth]{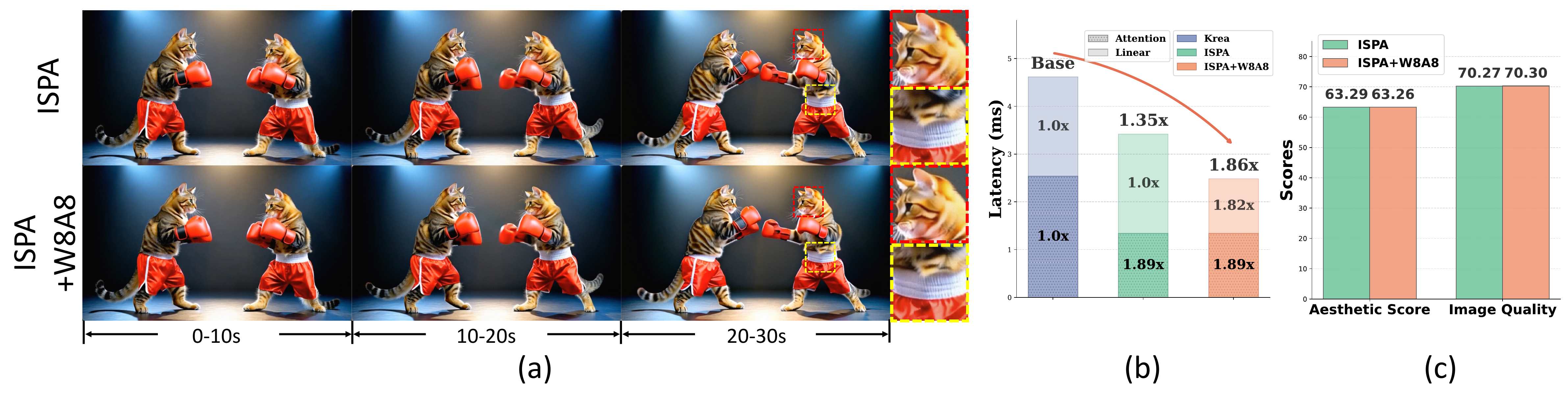} 
    \caption{Compatibility with Post-Training Quantization (PTQ). (a). Visual comparison between the standard ISPA and its quantized version (ISPA+W8A8). (b). Overall latency breakdown. (c). Quantitative stability analysis based on generation quality.}
    \label{fig:quant_efficiency}
\end{figure*}

\subsection{Further Analysis}
\noindent \textbf{Extension to Speech-to-Video (S2V) model.} 
To further validate the versatility of ISPA beyond text-to-video generation, we integrate it into the speech-to-video model (LiveAvatar). Speech-to-video synthesis is a demanding streaming task where the model must maintain strict identity consistency and precise lip-sync over hundreds of frames.
As illustrated in Figure~\ref{fig:s2v_liveavatar}, we compare ISPA against the full-cache baseline (LiveAvatar). It can be observed that ISPA successfully preserves the speaker’s identity and fine-grained facial expressions. Crucially, the mouth shapes (visemes) remain accurately synchronized with the input phonemes (e.g., /v/, /u:/). This indicates that our parametric update $\Delta W$ effectively "remembers" the specific facial dynamics of the speaker, confirming ISPA as a highly adaptable solution for diverse autoregressive architectures.

\noindent \textbf{Compatibility with Post-Training Quantization (PTQ).} 
While ISPA effectively mitigates the KV cache bottleneck in the attention mechanism, the computational load of linear projection layers remains a significant factor for end-to-end throughput. To address this, we explore the compatibility of ISPA with 8-bit weight and activation quantization (W8A8), aiming for a holistic acceleration of the Diffusion Transformer backbone. 
As illustrated in Fig.~\ref{fig:quant_efficiency} (b), the standard ISPA delivers a $1.35\times$ overall speedup by drastically optimizing the attention blocks (achieving 1.89$\times$ within the operator). By further integrating W8A8 quantization for the linear layers, we reach an aggregate speedup of \textbf{1.86$\times$} compared to the full-precision baseline. Importantly, our parametric update $\Delta W$ is derived through a stable, closed-form reconstruction, which inherently maintains the activation distributions. This property makes ISPA highly quantization-friendly. As shown in the qualitative results (Fig.~\ref{fig:quant_efficiency}(a)), ISPA+W8A8 exhibits no visual degradation or structural drifting. The quantitative scores for aesthetics and image quality (Fig.~\ref{fig:quant_efficiency}(c)) remain virtually unchanged after quantization. These results suggest that ISPA can serve as a robust foundation for ultra-efficient streaming systems.

\section{Related Works}

\noindent \textbf{Diffusion and Flow-Matching for Video Generation.} 
Recent advancements in video generation are primarily driven by scaling Diffusion Models and Rectified Flow to the temporal dimension. Building upon the success of Diffusion Transformers (DiT)~\cite{peebles2023scalable}, state-of-the-art models such as Sora, Latte~\cite{ma2024latte}, and MovieGen~\cite{polyak2024movie} have demonstrated remarkable capabilities in synthesizing high-fidelity videos. These frameworks typically employ 3D spatiotemporal attention to capture inter-frame dependencies. However, since the self-attention complexity grows quadratically with the number of tokens $\mathcal{O}(F^2)$, these holistic, offline generation paradigms are fundamentally limited by hardware memory, making the synthesis of long-form or high-resolution videos computationally prohibitive.

\noindent \textbf{Autoregressive (AR) Video Generation.} 
To circumvent the memory bottleneck of offline models, the Autoregressive (AR) paradigm~\cite{liu2025rolling,chen2025skyreels,team2026advancing,li2026train} has emerged as the standard for long-video synthesis. By partitioning a video into sequential chunks and generating them conditioned on previous frames, AR models~\cite{zhu2026causal,guo2026efficient,wu2025geometry,yi2025deep,yesiltepe2025infinity,cui2025self,wu2025pack,li2025stable,ma2026flow} theoretically enable infinite video extrapolation. A pioneer work is Causvid~\cite{yin2025slow}, which distills non-causal models into a autoregressive few-step model with DMD~\cite{yin2024improved,yin2024one} and trajectory distillation~\cite{song2023consistency,songimproved}.
The temporal condition is typically implemented via Key-Value (KV) cache injection, where the current chunk attends to a persistent memory bank of historical tokens. While effective, this strategy shifts the burden to memory capacity; as the video progresses, the linear accumulation of the KV cache eventually leads to memory saturation and significant inference latency spikes.

\noindent \textbf{KV Cache Compression in LLM.} 
KV cache compression has been extensively studied in Large Language Models (LLMs), where the main goal is to reduce the memory cost of long-context decoding while preserving language modeling quality. A common line of work~\cite{behnam2025rocketkv,liu2024minicache,li2024snapkv,tangquest,xiao2023efficient,hooper2025squeezed,wangprefixkv} exploits the sparsity structure of attention. StreamingLLM~\cite{xiao2023efficient} and related studies observe that a small number of initial tokens, known as attention sinks~\cite{xiao2023efficient,guattention,yu2024unveiling}, receive consistently high attention and are important for stabilizing streaming inference; therefore, they keep these sink tokens together with recent tokens while evicting the middle context. Other methods estimate token importance from attention statistics. For example, H2O~\cite{zhang2023h2o} retains heavy-hitter tokens that accumulate large attention scores, while SnapKV~\cite{li2024snapkv} uses an observation window to identify task-relevant KV tokens before generation. More recent methods further refine this idea with query-aware or multi-stage compression, such as QUEST~\cite{tangquest} and RocketKV~\cite{behnam2025rocketkv}, which adapt KV selection according to the current decoding query or combine coarse and fine pruning stages. 

Despite their effectiveness in LLMs, these methods still manage KV cache as an external token memory by deciding which cached states to preserve or discard. This paradigm is fragile for autoregressive video generation, where historical KV tokens encode dense visual states and long-range temporal consistency. ISPA therefore departs from token-level cache management by absorbing part of the historical context into instance-specific weight modulation.

\section{Conclusion}
In this work, we rethink the challenge of long-range dependency in video generation and propose ISPA. Rather than viewing KV cache compression merely as a problem of selecting or discarding tokens, we treat the growing KV cache as an external memory of the streaming model and explore how part of this memory can be internalized into the model itself. Through instance-specific parametric absorption, ISPA consolidates historical context into lightweight weight modulation at test time, allowing selected layers to operate with local attention while preserving long-range temporal coherence.

Enabled by our Decomposable Attention mechanism and closed-form adaptation, this process provides a persistent, constant-memory representation of video history without gradient-based optimization. Extensive evaluations across DiT-based architectures ranging from 1.3B to 14B parameters demonstrate that ISPA removes up to 50\% of the KV cache with near-lossless visual fidelity, while its compatibility with quantization delivers an aggregate 1.86× inference speedup. Beyond these empirical gains, we hope this work encourages the community to look beyond external token-level cache management and further explore parametric memory consolidation as a general route toward efficient streaming generative models.



%
%
\bibliographystyle{splncs04}
\bibliography{main}
\end{document}